\documentclass[sigconf]{acmart}
\AtBeginDocument{%
  \providecommand\BibTeX{{%
    \normalfont B\kern-0.5em{\scshape i\kern-0.25em b}\kern-0.8em\TeX}}}

\copyrightyear{2021}
\acmYear{2021}
\setcopyright{acmcopyright}\acmConference[MM '21]{Proceedings of the 29th ACM International Conference on Multimedia}{October 20--24, 2021}{Virtual Event, China}
\acmBooktitle{Proceedings of the 29th ACM International Conference on Multimedia (MM '21), October 20--24, 2021, Virtual Event, China}
\acmPrice{15.00}
\acmDOI{10.1145/3474085.3475458}
\acmISBN{978-1-4503-8651-7/21/10}

\acmSubmissionID{mfp1623}

\usepackage{amsmath}
\usepackage{siunitx}
\usepackage{booktabs}
\usepackage{array}
\usepackage{multirow}
\usepackage{subfigure}
\usepackage{algorithm}
\usepackage{algpseudocode}
\usepackage[utf8]{inputenc}
\usepackage{graphicx}
\usepackage{epstopdf}
\usepackage{listings}
\usepackage{marvosym}
\usepackage{balance}
\def\ie{{\textit{i.e.}}}
\def\eg{{\textit{e.g.}}}
\def\etc{{\textit{etc.}}}

\def\x{{\mathbf x}}

\def\f{{\mathbf f}}

\def\h{{\mathbf h}}

\def\F{{\mathbf F}}

\def\H{{\mathbf H}}
\def\M{{\mathbf M}}
\def\N{{\mathbf N}}

\def\P{{\mathbf P}}

\begin{document}
\fancyhead{}

%%
%% The "title" command has an optional parameter,
%% allowing the author to define a "short title" to be used in page headers.
\title{Self-Contrastive Learning with Hard Negative Sampling \\ for Self-supervised Point Cloud Learning}

%%
%% The "author" command and its associated commands are used to define
%% the authors and their affiliations.
%% Of note is the shared affiliation of the first two authors, and the
%% "authornote" and "authornotemark" commands
%% used to denote shared contribution to the research.

%%
%% By default, the full list of authors will be used in the page
%% headers. Often, this list is too long, and will overlap
%% other information printed in the page headers. This command allows
%% the author to define a more concise list
%% of authors' names for this purpose.
\author{Bi'an Du}
\authornote{Both authors contributed equally to this research.}
\email{scncdba@pku.edu.cn}
\author{Xiang Gao}
\authornotemark[1]
\email{gyshgx868@pku.edu.cn}
\affiliation{%
  \institution{Wangxuan Institute of Computer Technology, Peking University}
  \streetaddress{No. 128 Zhongguancun North Street}
  \city{Beijing}
  \country{China}
}

\author{Wei Hu}
\authornote{This work was supported by National Natural Science Foundation of China (61972009). Corresponding author: Wei Hu (forhuwei@pku.edu.cn).}
\email{forhuwei@pku.edu.cn}
\affiliation{%
  \institution{Wangxuan Institute of Computer Technology, Peking University}
  \streetaddress{No. 128 Zhongguancun North Street}
  \city{Beijing}
  \country{China}}

\author{Xin Li}
\email{xin.li@mail.wvu.edu}
\affiliation{%
  \institution{Lane Department of Computer Science and Electrical Engineering, West Virginia University}
  \streetaddress{PO Box 6109}
  \city{Morgantown, WV}
  \country{USA}}

\renewcommand{\shortauthors}{Bi'an Du and Xiang Gao, et al.}

\begin{abstract}
   Point clouds have attracted increasing attention. Significant progress
has been made in methods for point cloud analysis, which often requires
costly human annotation as supervision. To address this issue,
we propose a novel self-contrastive learning for self-supervised point
cloud representation learning, aiming to capture both local geometric
patterns and nonlocal semantic primitives based on the nonlocal
self-similarity of point clouds. The contributions are two-fold: on
the one hand, instead of contrasting among different point clouds as
commonly employed in contrastive learning, we exploit self-similar
point cloud patches within a single point cloud as positive samples
and otherwise negative ones to facilitate the task of contrastive learning.
On the other hand, we actively learn hard negative samples
that are close to positive samples for discriminative
feature learning. Experimental
results show that the proposed method achieves state-of-the-art
performance on widely used benchmark datasets for self-supervised
point cloud segmentation and transfer learning for classification.
\end{abstract}

%%
%% The code below is generated by the tool at http://dl.acm.org/ccs.cfm.
%% Please copy and paste the code instead of the example below.
%%



%%
%% Keywords. The author(s) should pick words that accurately describe
%% the work being presented. Separate the keywords with commas.

%% A "teaser" image appears between the author and affiliation
%% information and the body of the document, and typically spans the
%% page.

%%
%% This command processes the author and affiliation and title
%% information and builds the first part of the formatted document.
\maketitle

\section{Introduction}
% background
3D point clouds serve as an efficient representation of 3D objects or natural scenes, which consist of irregularly sampled 3D points associated with multiple modalities, leading to a wide range of applications such as autonomous driving, robotics and immersive tele-presence.
% While point clouds consist of discrete 3D points
% irregularly sampled from continuous surfaces that cannot be directly addressed by Convolutional Neural Networks (CNNs)
Recent advances in geometric deep learning \cite{bronstein2017geometric} have shown their success in the representation learning of irregular point clouds.
However, most methods are trained in a (semi-)supervised fashion, requiring a large amount of labeled data to learn adequate feature representations.
This limits the wide applicability of point clouds, especially for large-scale graphs.
Hence, it is desirable to learn the feature representations of point clouds in a self-supervised fashion.

\begin{figure}
  \centering
  \includegraphics[width=\columnwidth]{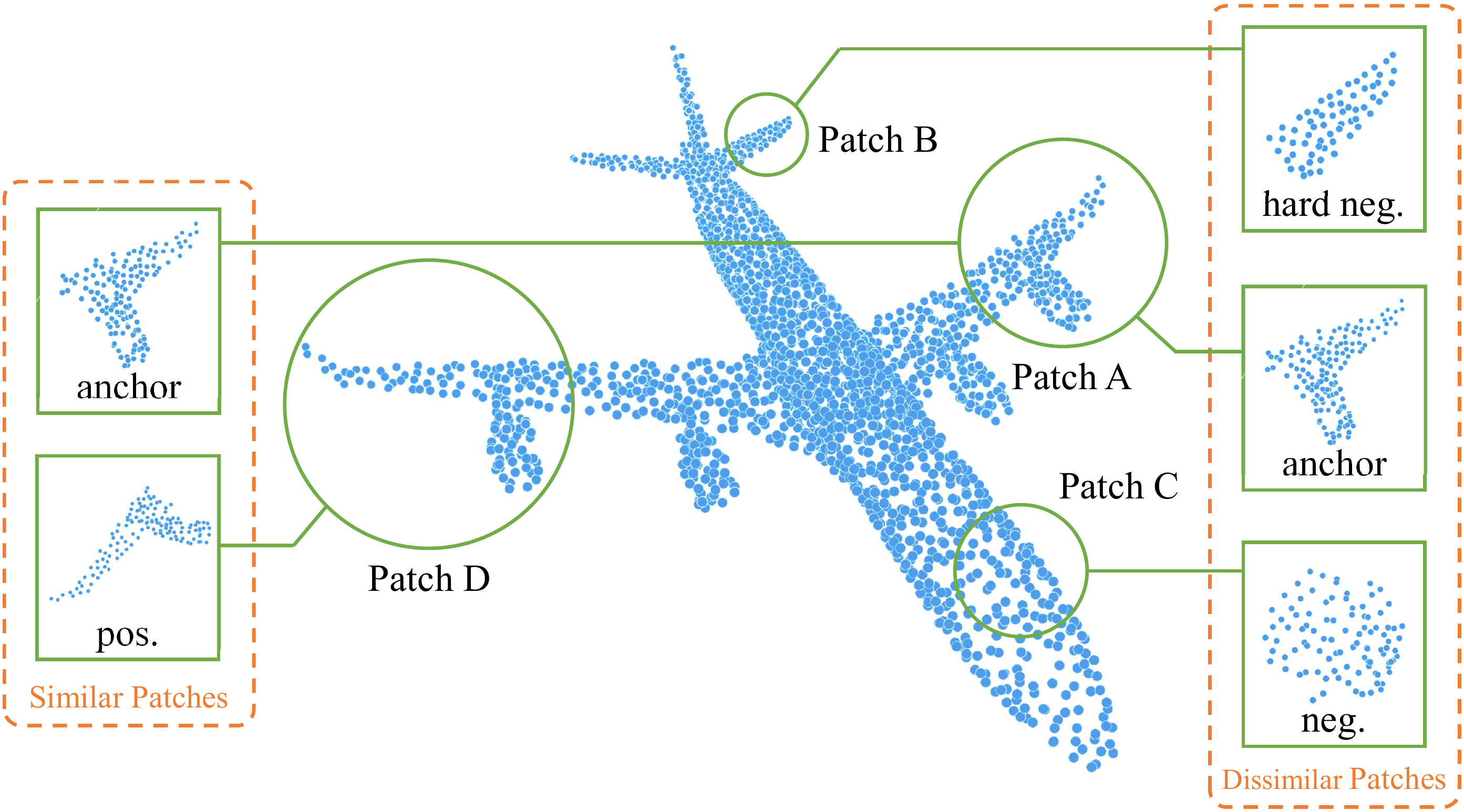}
  \vspace{-0.25in}
  \caption{An illustration of the proposed self-contrastive learning for self-supervised point cloud representation learning. Patch A is the anchor; \textit{pos.} and \textit{neg.} denote positive (\eg, patch D) and negative (\eg, patch B or patch C) samples, respectively. Note that patch B is the {\it hard} negative sample because of its comparative similarity to the anchor.}
  \vspace{-0.2in}
  \label{fig:teaser}
\end{figure}

Several attempts have been made for self-supervised representation learning on point clouds.
These approaches are mainly based on reconstruction \cite{deng2018ppf,yang2018foldingnet,liu2019l2g,han2019multi,zhao20193d,sauder2019self,gao2020graphter} or generation \cite{achlioptas2018learning,valsesia2018learning,gadelha2018multiresolution,li2019point,sun2020pointgrow}.
% The former aims at training an encoder to learn feature representations by reconstructing the input data via a decoder;
% while the latter attempts to learn the feature representation by training Generative Adversarial Networks (GANs) \cite{goodfellow2014generative} or Variational Auto-Encoders (VAEs) \cite{kingma2014auto} to generate 3D point clouds.
% Also, many approaches have sought to explore semantic information \cite{zhang2019unsupervised,rao2020global} or equivariant representations \cite{} for point cloud learning.
% These methods have demonstrated to be effective in capturing structural and low-level information of point clouds, but usually fail to learn high-level semantic information \cite{rao2020global}.
Besides, contrastive learning instantiates a family of self-supervised methods \cite{hjelm2018learning,bachman2019learning,chen2020simple,he2020momentum,hu2021adco,sanghi2020info3d,xie2020pointcontrast}, which often maximizes the agreements between the augmented views of the same image in an embedding feature space, while avoiding the mode collapse of the embedded features by maximizing the disagreements between negative examples constructed from different images.
This paradigm has been extended to point cloud learning \cite{zhang2019unsupervised,sanghi2020info3d,xie2020pointcontrast}, which either contrast among various point clouds or different projected views of the input point cloud for representation learning.
Under no supervision, the success of previous methods depends on informative sampling of positive and negative pairs, which often resort to manual augmentation (\eg, projections) or additional data (\eg, different point clouds for training).

To this end, we propose a novel framework for self-supervised point cloud representation learning via {\em self-contrastive learning}, which actively {\it learns} positive and negative samples from the input {\it single} point cloud, motivated by the nonlocal self-similar property of point clouds.
The key observation is that, as a representation of 3D objects or scenes, a point cloud usually exhibits nonlocal self-similarity, \ie, similar or even the same local geometry after affine transformations, such as symmetric engines and airfoils of an airplane as shown in Fig.~\ref{fig:teaser}.
Based on this observation, we propose to learn self-similar point cloud patches as positive samples and otherwise negative ones, without resorting to other point clouds or additional projections. Such self-supervised approach marks the first significant departure from the standard practice of contrastive learning.
Moreover, both local geometric patterns and nonlocal semantic primitives are jointly exploited by self-contrastive learning to improve the robustness against noise and missing data.
As negative sampling is crucial for discriminative feature learning \cite{yang2020understanding, 2020Contrastive, wu2020conditional}, we learn {\it hard} negative samples that are close to positive samples in the representation space for more expressive representations.
Conditioned on each anchor patch, the corresponding hard negative samples are inferred based on the degree of self-similarity with the anchor.

Specifically, given an input point cloud, we first train a similarity-learning network to infer the similarity between each pair of point cloud patches.
To train the network, we apply rotations to each anchor patch to generate similar samples and employ another different patch in the point cloud as a dissimilar sample as input to learn patch similarity.
Then, given patch pairs in the input, we discriminate if they are similar or dissimilar pairs according to the estimated of the similarity-learning network.
Similar patches are treated as positive samples while dissimilar ones are negative samples.
Furthermore, we actively sample {\it hard} negative patches---negatives with comparatively large similarity to the anchor patch, where the similarity is measured by the cosine similarity between the features of patch pairs inferred from the similarity-learning network, aiming at more discriminative contrastive learning.
Since the contrastive network is randomly initialized, naively employing hard negative samples may fall into an unsatisfactory local minimum. To avoid this, we first choose the whole negative samples for contrastive learning, and then perform linear annealing \cite{wu2020conditional} on the thresholds of self-similarity to choose hard negative samples.

Our main contributions are summarized as follows.

  1) We propose a novel self-supervised framework for point cloud representation learning by contrasting patches within each point cloud, aiming at exploiting the {\bf nonlocal self-similarity} of point clouds. Unlike previous works resorting to different point clouds or projections, we advocate the use of self-similar patches within a single point cloud as positive samples and otherwise negative samples for contrastive learning.

  2) To close the gap between supervised and unsupervised learning, we have developed an effective {\bf hard negative sampling} method for learning discriminative features. We exploit the nonlocal self-similarity of point clouds to determine hard negative sampling and employ a linear annealing strategy to dynamically choose the hard negative samples for contrastive learning to avoid falling into an unsatisfactory local minimum.
%   Conceptually similar to recent work on graph representation learning \cite{yang2020understanding}, we for the first time attempt to demonstrate that negative sampling is as important as positive sampling for point cloud segmentation.

  3) Experimental results show that the proposed model {\bf outperforms current state-of-the-art methods} in point cloud segmentation and transfer learning for classification on widely used benchmark datasets, and validate the proposed informative hard negative sampling.
 \vspace{-0.1in}

\section{Related Work}
\begin{figure*}[t]
  \centering
  \includegraphics[width=\textwidth]{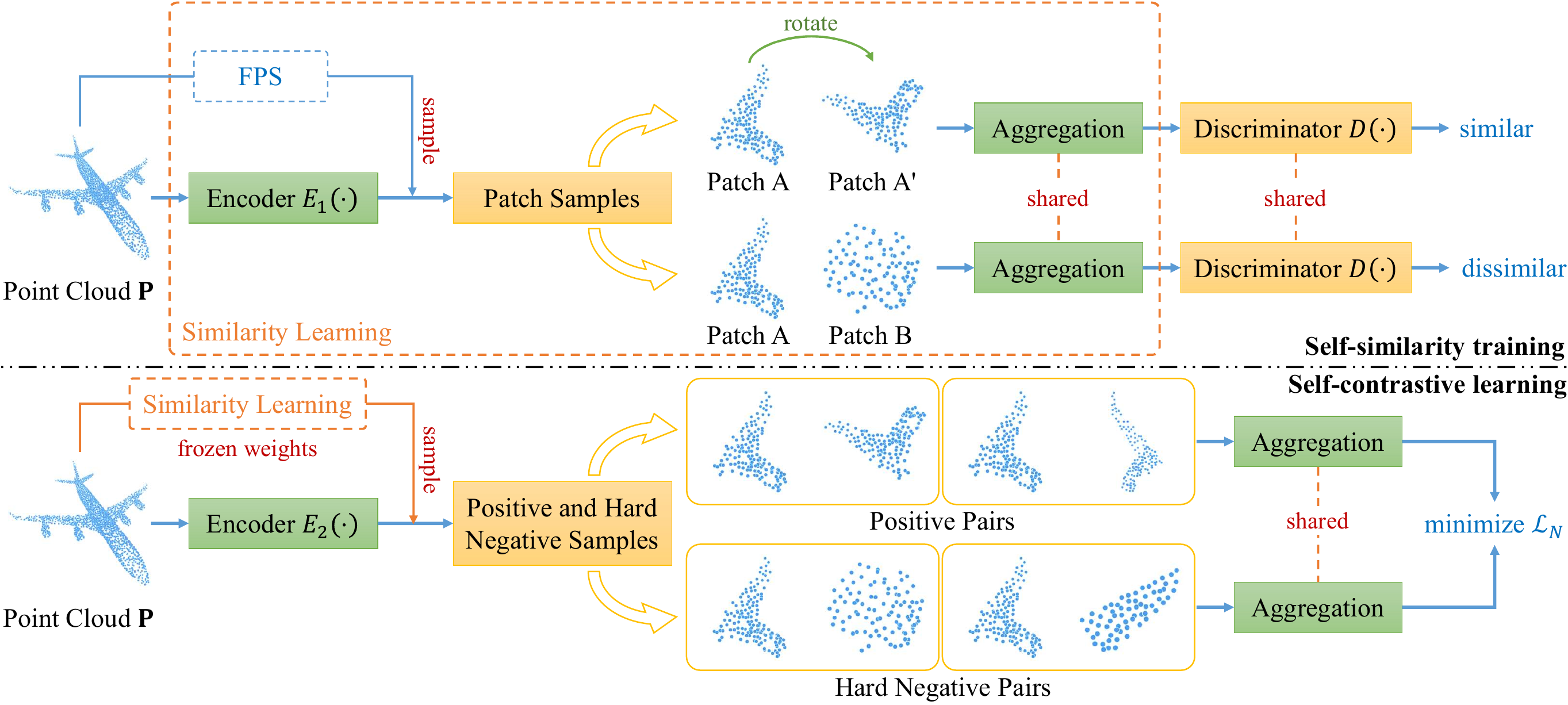}
  \vspace{-0.25in}
  \caption{The architecture of the proposed model for self-supervised point cloud representation learning, which consists of self-similarity training and self-contrastive learning.}
  \vspace{-0.1in}
  \label{fig:architecture}
\end{figure*}

\subsection{Self-supervised Representation Learning}

\subsubsection{Auto-Encoders and Generative Adversarial Networks}

% Auto-Encoders (AEs) and Generative Adversarial Networks (GANs) are two most representative methods for unsupervised representation learning.
Auto-encoders (AEs) aim to train an encoder to learn feature representations by reconstructing the input data via a decoder \cite{vincent2008extracting,rifai2011contractive,hinton2011transforming,kingma2014auto}.
The idea is based on that feature representations should contain sufficient information to reconstruct the input data.
A plethora of approaches have been proposed to learn unsupervised feature representations via AEs, including denoising AEs \cite{vincent2008extracting}, contrastive AEs \cite{rifai2011contractive}, transforming AEs \cite{hinton2011transforming}, variational AEs (VAEs) \cite{kingma2014auto}, \etc.
In addition to AEs, Generative Adversarial Networks (GANs) \cite{goodfellow2014generative,donahue2017adversarial,dumoulin2017adversarially} extract feature representations in an unsupervised fashion by generating data from input noise via a pair of generator and discriminator.
Recent approaches have shown great potential of GANs in providing expressive feature representations by generating images \cite{goodfellow2014generative,donahue2017adversarial,dumoulin2017adversarially}, point clouds \cite{achlioptas2018learning,valsesia2018learning,li2019point}, or graphs \cite{li2018learning,you2018graphrnn,de2018molgan,bojchevski2018netgan}.

\vspace{-0.1in}
\subsubsection{Contrastive Representation Learning}

Contrastive learning instantiates a wide range of self-supervised methods \cite{hjelm2018learning,bachman2019learning,chen2020simple,he2020momentum,hu2021adco,sanghi2020info3d,xie2020pointcontrast}, which aims to train an encoder to be \textit{contrastive} between the feature representations of positive samples and negative samples.
Among them, Deep InfoMax \cite{hjelm2018learning} first proposed to maximize the mutual information between a local patch and its global context through a contrastive learning framework.
% AMDIM \cite{bachman2019learning} maximizes the mutual information between feature representations of multiple views of a shared context.
SimCLR \cite{chen2020simple} proposed a simple framework for contrastive learning, which requires a large batch size to achieve superiority performance.
MoCo \cite{he2020momentum} improves the efficiency of contrastive learning by introducing a queue to store feature representations, which reduces the memory cost for negative samples.
AdCo \cite{hu2021adco} presents an adversarial approach to demonstrate the negative pairs of examples can be directly trained end-to-end together with the backbone network so that the contrastive model can be learned more efficiently as a whole.
Recently, Info3D \cite{sanghi2020info3d} proposed to extend the InfoMax and contrastive learning framework on 3D objects, which maximizes the mutual information between 3D objects and their ``chunks'' to learn representations.
PointContrast \cite{xie2020pointcontrast} proposed a unified framework of contrastive loss for representation learning on 3D scenes.
In addition, hard negative sampling has shown to benefit contrastive learning in \cite{wu2020conditional,kalantidis2020hard}.
Robinson \textit{et al.} \cite{2020Contrastive} proposed a new conditional distribution for sampling negative samples to distinguish false-negative samples with the same context information as the anchor from hard negative samples.
Wu \textit{et al.} \cite{wu2020conditional} proposed a negative sample selection method with upper and lower bounds, which calculates a certain distance between the samples and then takes the samples between the upper and lower bounds as negative samples for learning.

\vspace{-0.1in}
\subsubsection{Transformation Equivariant Representation Learning}

Another family of self-supervised approaches attempt to learn transformation equivariant representations, which has been advocated in Hinton's seminal work on learning transformation capsules \cite{hinton2011transforming}.
Following this, many approaches have been proposed to learn transformation equivariant representations \cite{gens2014deep,dieleman2015rotation,dieleman2016exploiting,cohen2016group,lenssen2018group}.
To generalize to generic transformations, Auto-Encoding Transformation (AET) was proposed in \cite{zhang2019aet} to learn unsupervised feature representations by estimating transformations from the learned feature representations of both the original and transformed images, which was further extended in \cite{qi2019avt,wang2020transformation,gao2021self}.
Another extension of AET, named GraphTER \cite{gao2020graphter}, was introduced to graph-structured data formalized by auto-encoding node-wise transformations in an unsupervised manner.
The self-attention mechanism was specifically introduced in \cite{fuchs2020se} for 3D point cloud data, which adheres to equivariance constraints, improving robustness to nuisance transformations.

\vspace{-0.1in}
\subsection{Deep Point Cloud Learning}

\subsubsection{Deep Learning on Point Clouds}
Deep learning on 3D point clouds has attracted increasing attention in recent years.
Many approaches have been proposed to address various tasks on point clouds, such as 3D point cloud classification and segmentation \cite{qi2017pointnet,qi2017pointnet++,guerrero2018pcpnet,li2018pointcnn,wu2019pointconv,liu2019relation,te2018rgcnn,zhang2018graph,wang2018local,wang2019dynamic}.
Among them, one pioneer method PointNet \cite{qi2017pointnet} proposed to learn the features of each point independently, while PointNet++ \cite{qi2017pointnet++} introduces a hierarchical architecture that applies PointNet on a nested partitioning of the input point set to extract local structures.
% PointCNN \cite{li2018pointcnn} generalizes CNN by leveraging spatially local correlation from data represented in point clouds.
% PCNN \cite{atzmon2018point} defines convolution of functions over point clouds, aiming to translate the volumetric convolution to arbitrary point clouds using extension and restriction operators.
% Relation-Shape CNN (RS-CNN) \cite{liu2019relation} extends regular grid CNN to irregular configuration for point cloud analysis.
% PointConv \cite{wu2019pointconv} trains multi-layer perceptrons on local point coordinates to approximate continuous weight and density functions in convolutional filters, which makes point sets permutation-invariant and translation-invariant.
Local structures have also been exploited by methods such as PCNN \cite{atzmon2018point}, PointCNN \cite{li2018pointcnn}, PointConv \cite{wu2019pointconv}, and Relation-Shape CNN \cite{liu2019relation} to further improve the quality of point cloud representation learning.
In addition, Graph Convolutional Neural Networks (GCNNs) have also been applied to point clouds by constructing a $k$-nearest-neighbor ($k$NN) graph or a complete graph to learn feature representations \cite{te2018rgcnn,zhang2018graph,wang2018local,wang2019dynamic}.
% These works have promoted the development of deep learning on point clouds, and also pointed a new route to learn adequate feature representations in an unsupervised fashion.
\vspace{-0.1in}

\subsubsection{Self-supervised Point Cloud Learning}
Recently, many approaches have sought to explore semantic information for unsupervised point cloud learning.
In \cite{zhang2019unsupervised}, a method was proposed to learn unsupervised semantic features for point clouds by solving the pretext tasks of part contrasting and object clustering.
\cite{sauder2019self} attempted to reconstruct point clouds whose parts have been randomly rearranged to capture semantic properties of the point cloud.
These representations are learned by inferring the relationship among parts, while the global information is not fully exploited.
To tackle this issue, a recent work \cite{rao2020global} proposed to learn point cloud representations by bidirectional reasoning between the local structures and the global shape.
However, the semantic nonlocal context information within each point cloud is not fully exploited yet in previous works.

% \section{The Proposed Formulation}
% \label{sec:formulation}
% \input{3_formulation}

% \vspace{-0.1in}
\section{The Proposed Method}
In this section, we elaborate on the proposed self-supervised point cloud representation learning. We start with an overview of the key ideas, and then present our self-contrastive learning with hard negative sampling from nonlocal self-similarity of point clouds.
Furthermore, we provide analysis about the relationship of our method to  self-attention \cite{zhao2020point} and point cloud transformer \cite{guo2020pct}.

\vspace{-0.1in}
\subsection{Overview}
Given an input point cloud $\P \in \mathbb{R}^{N \times 3}$, we aim at learning an effective feature extractor $\mathcal{E}: \P \mapsto f(\P)$ that infers the feature $f(\P)$ of $\P$ in a self-supervised fashion.
As illustrated in Fig.~\ref{fig:architecture}, our framework consists of the following three procedures:
\begin{itemize}
    \item \textbf{Self-similarity training.} We train a similarity-learning network to learn the similarity between each pair of point cloud patches in $\P$.
    Specifically, for each anchor patch, we generate pairs of similar patches by rotation of the anchor and employ another different patch in $\P$ as a dissimilar sample, which serve as the training data to learn the patch similarity.

    \item \textbf{Positive and Hard negative sampling.} We employ the trained similarity-learning network to learn positive and negative samples, depending on if they are similar or dissimilar pairs.
    Meanwhile, we propose to determine {\it hard} negative samples by the degree of self-similarity in the feature space inferred from the similarity-learning network. Such hard negative examples will play an important role in facilitating better and faster contrastive learning \cite{kalantidis2020hard}.

    \item \textbf{Self-contrastive learning.} Based on the learned positive and hard negative samples, we perform self-contrastive learning over $\P$. In particular, we adopt a linear schedule in deterministic annealing for hard negative samples to avoid falling into a local minimum during self-contrastive learning.
\end{itemize}

\vspace{-0.1in}
\subsection{Nonlocal Self-similarity in Point Clouds}
\label{subsec:nonlocal}

Point clouds often exhibit nonlocal self-similarity, as illustrated in Fig.~\ref{fig:teaser}. Unlike image patches, two point cloud patches are similar if they can be connected by 3D geometric transformations (rotation, reflection, translation, \etc).
Since point clouds characterize 3D shapes, there is often stronger self-similarity in point clouds than that in images, \eg, six sides of a cube are similar to each other, left and right sides of many objects including human faces observe bilateral symmetry, and circular symmetry can be widely observed for vases, balls, and so on.

This key observation inspires our self-contrastive learning, where self-similar patches serve as positive samples and dissimilar patches become negative samples.
To fully exploit the nonlocal self-similarity, it is crucial to provide a meaningful definition of nonlocal self-similarity for point clouds. Based on the heuristic that two patches in a point cloud are self-similar if their distance is small after some 3D geometric transformation, there are many optimization-based registration methods that use optimization strategies to estimate the transformation matrix \cite{cheng2018registration,le2019sdrsac,yang2019polynomial}, and then judge the similarity of two point sets by exploring the properties of the transformation matrix. However, these methods usually require high computational cost for complex matching strategies and easily fall into a local optimal value.
Different from explicitly calculating the shape similarity between patches using optimization strategies, we consider projecting different patches into the same similarity feature space via a similarity-learning network. Next, we measure the similarity between patches by calculating the cosine similarity of the corresponding feature vectors.

\textbf{Definition 1.} Given a point cloud $\P$, two patches $\M \in \P$ and $\N \in \P$ are self-similar if
\begin{equation}
    \Phi[g(\M),g(\N)] < \epsilon,
\end{equation}
where $g(\cdot)$ denotes the feature of a patch learned from a network, and $\Phi[\cdot,\cdot]$ is a metric to measure the similarity between the features of patch pairs, such as the cosine similarity. $\epsilon >0$ is a small threshold.

\subsection{Self-similarity Training}
\label{subsec:self_similar_train}

Based on \textbf{Definition 1}, we train a similarity-learning network to learn the self-similarity between point cloud patch pairs.
As shown in the top half of Fig.~\ref{fig:architecture}, given a point cloud $\P=\{\x_i\}_{i=1}^{N}$, we choose patch centers by iterative farthest point sampling (FPS) \cite{moenning2003fast}, leading to a subset of $M$ points $\mathbf{C}=\left\{\x_i\right\}_{i=1}^{M}$.
We thus construct a set of patches by
\begin{equation}
  \mathcal{P}_i=\left\{\x_j | \x_j \in \mathcal{N}(\x_i) \cup \x_i \right\}, i=1...,M,
\end{equation}
where $\mathcal{N}(\cdot)$ denotes the $k$ nearest neighbors of the center point $\x_i$ (\textit{i.e.}, $|\mathcal{P}_i|=k+1$).

To learn the features of these patches, we first introduce an encoder $E_1(\cdot)$ that maps the point cloud $\P$ to the feature space,
\begin{equation}
  \F=E_1(\P),
\end{equation}
where $\F=[\f_1,...,\f_N]^\top\in\mathbb{R}^{N \times F}$ is the feature matrix of the point cloud with $F$ output feature channels, with $\f_i$ as the feature of the $i$-th point.
Specifically, we employ DGCNN \cite{wang2019dynamic} as the encoder.
Then each constructed patch is represented by the concatenation of the feature of each point in the patch, \ie, $\F_i \in \mathbb{R}^{(k+1)\times F}$, $i=1,...,M$.
% By using the sampled $M$ patches $\mathcal{P}_i$, we represent the patch in feature space as
% \begin{equation}
%   \mathcal{F}_i = \left\{\f_j | \forall j, \x_j \in \mathcal{P}_i \right\}, i=1,...,M,
%   \label{eq:feature_patch_sample}
% \end{equation}
% where $|\mathcal{F}_i|=k+1$ denotes the feature dimension.

In order to learn the nonlocal self-similarity of patches, given an anchor patch $\mathcal{P}_i$, we first construct a similar sample by rotating $\mathcal{P}_i$ with a random degree, which admits a feature vector representation $\widetilde{\F}_i$. Meanwhile, we randomly sample another patch $\mathcal{P}_j, i \neq j$ as a dissimilar patch to $\mathcal{P}_i$.
Then we summarize the patch representation by an aggregation operator $G_1:\mathbb{R}^{(k+1) \times F} \mapsto \mathbb{R}^F$ for the subsequent discrimination.

% a patch in $\P$ is constructed as a set of $k+1$ points, which consists of a centering point and its $k$-nearest neighbors.
% We choose patch centers  as patch centers, on which we apply a K-nearest neighbor(kNN) search.
% Here each point contains both coordinates and normals, \textit{i.e.}, $x_i \in \mathbb{R}^{6},\space \forall i \in  \left\{1,2,...,n \right\}.$

% Next, we define an encoder $E_1: M\longmapsto S$ that maps input patches from $\mathbb{R}^{m*d_i}$ into the latent space $S \in \mathbb{R}^{m*d_s}$ with $d_s \gg d_i$. For each patch, the learned feature $\left\{s_1,s_2,...,s_m\right\} \in \mathbb{R}^{m*d_s}$ are fed into a GCN layer $\mathcal{A}$, where we aggregate them into a shape feature $A_i \in \mathbb{R}^{d_{s^'}}$.

% By applying different transformation on the same patch and choosing different patches, we get similar samples and dissimilar samples for the Discriminator $D(\cdot)$, which is used to perform the binary classification task---1 for similar samples and 0 for dissimilar samples. Specifically, we randomly rotate patch $\M$ with three rotation parameters of X-axis, Y-axis and Z-axis in the range $\left[0^{\circ}, 180^{\circ} \right]$ to get the similar sample $\M^{+}$, and we perform patch index permutation to get the dissimilar sample $\N$.

Next, we employ a discriminator $D: \left(\mathbb{R}^{F}, \mathbb{R}^{F}\right) \mapsto \mathbb{R}$, where $D(G_1(\F_i),G_1(\F_j))$ provides the similarity score assigned to the summarized representations of patch pair $(\mathcal{P}_i,\mathcal{P}_j)$.
A higher score corresponds to a more similar patch pair.
The similarity-learning network is trained by minimizing
\begin{equation}
\begin{split}
  \mathcal{L}_S = -\frac{1}{A+B}\bigg( & \sum_{i=1}^{A} \log D\left(G_1(\F_i),G_1(\widetilde{\F}_i)\right) \\
  & + \sum_{i=1}^{B} \log \Big(1-D\left(G_1(\F_i),G_1(\F_j)\right)\Big)\bigg),
\end{split}
\end{equation}
where $A$ and $B$ are the numbers of sampled similar and dissimilar patch pairs, respectively.

\subsection{Positive and Hard Negative Sampling}

% Given an anchor patch $\mathcal{P}_i$, negative examples are sampled i.i.d. from the marginal distribution of $\mathcal{P}_i$, \ie, $p(\mathcal{P}_i)$.
% Instead, we extract negative samples from a distribution $q^{-}(\mathcal{P}_j|\mathcal{P}_i)$ conditional on an anchor patch $\mathcal{P}_i \sim p(\mathcal{P}_i)$.
Two key ingredients in contrastive learning are definitions of  similar (positive) pairs and dissimilar (negative) pairs of data points.
Having trained the similarity-learning network for self-similarity learning, we proceed with inferring negative samples from the network.
Unlike most previous works where negative samples are manually chosen, we actively learn negative sampling conditional on each anchor patch.
Since we have no supervision in unsupervised contrastive learning, we opt to learn negative samples by measuring the similarity with respect to the anchor patch in the feature space.
In particular, we calculate the cosine similarity of the respective features of two patches inferred from the trained similarity-learning network, \ie,
%\begin{equation}
%  s(\mathcal{P}_i, \mathcal{P}_j) = \bigg| \frac{G_1(\F_i)^{\top}G_1(\F_j)}{\|G_1(\F_i)\|_2\cdot\|G_1(\F_j)\|_2} \bigg|,
%\end{equation}
\begin{equation}
  s(\mathcal{P}_i, \mathcal{P}_j) =  \frac{|G_1(\F_i)^{\top}G_1(\F_j)|}{\|G_1(\F_i)\|_2\cdot\|G_1(\F_j)\|_2} ,
\end{equation}

Furthermore, we propose to extract {\it hard} negative samples for discriminative feature learning inspired by recent works \cite{kalantidis2020hard,wu2020conditional}.
The level of ``hardness'' for negative samples is dependent on the similarity with respect to the anchor patch, \ie, more similar patches serve as ``harder'' negative samples.
To implement this idea, we choose a closed interval $\mathcal{B} = [b_l, b_u]$ defined by two thresholds for hard negative samples, where $b_l$ and $b_u$ are the lower and upper similarity thresholds, respectively.
% $b_u$ should be greater than the expected similarity with respect to $p(\mathcal{P}_i)$.
The negative sampling for each anchor patch $\mathcal{P}_i$ varies as the sampling thresholds are conditional on the anchor in the point cloud data.

Regarding positive sampling, we dilate each anchor patch as a positive example like in dilated point convolution \cite{2020Dilated}. Specifically, as anchor patch $\mathcal{P}_i$ is constructed via a $k$-nearest-neighbor search, we first compute the $k \times d$ nearest neighbors of the centering point ($d>1$ is an integer), and then select every $d$-th neighbor to acquire a dilated patch (with increased receptive field) as the positive sample $\mathcal{P}^+_i$. In our current experiments, we set $d=2$.

\subsection{Self-contrastive Learning}

Based on the learned positive and hard negative samples, we perform self-contrastive learning over $\P$.
We first learn the point-wise features of $\P$ through a representation encoder $E_2(\cdot)$,
\begin{equation}
  \H=E_2(\P),
\end{equation}
where $\H=[\h_n,...,\h_N]^{\top} \in \mathbb{R}^{N \times H}$ denotes the feature matrix of $\P$ with $\H$ output feature channels, with $\h_i$ as the feature of the $i$-th point.
The point-wise features will be transferred to downstream tasks, such as point cloud classification and segmentation.
We have employed DGCNN \cite{wang2019dynamic} as the encoder in our current implementation.
Each constructed patch is then represented by the concatenation of encoded features, \ie, $\H_i\in\mathbb{R}^{(k+1)\times H}, i=1,...,M$.

Like the similarity-learning network, we also employ an aggregation operator $G_2:\mathbb{R}^{(k+1)\times H} \mapsto \mathbb{R}^{H}$ to acquire the patch representations.
The contrastive learning network is then trained by minimizing the InfoNCE \cite{oord2018representation} contrastive loss,
\begin{equation}
  \mathcal{L}_N = -\log \frac{f(\H_i,\H_i^{+})}{f(\H_i,\H_i^{+})+\sum_{s(\mathcal{P}_i,\mathcal{P}_j)\in[b_l,b_u]}f(\H_i,\H_j)},
  \label{eq:info_nce}
\end{equation}
where $s(\mathcal{P}_i,\mathcal{P}_j)\in[b_l,b_u]$ means $\mathcal{P}_j$ is a hard negative sample of $\mathcal{P}_i$. $\H_i^{+}$ and $\H_j$ denote the feature map of the positive and negative patches, respectively.     $f(\H_i,\H_j)=\exp\left\{G_2(\H_i)^{\top}G_2(\H_j)/\tau\right\}$, where $\tau$ is a positive real number denoting the temperature parameter.

Naively using negative samples may collapse to poor local minimum as discussed in previous works \cite{wu2020conditional,kalantidis2020hard}.
Instead, we adaptively perform negative sampling by choosing the thresholds $b_l$ and $b_u$ with a linear annealing policy as in \cite{wu2020conditional}.
Specifically, we opt to reduce the value of $b_u$ by a step size linearly and increase the value of $b_l$ by another step size linearly for every fixed number of epochs after several training epochs of using the whole negative samples, thus selecting more difficult negatives.

\begin{table}[t]
\small
\caption{Classification results in transfer learning from ShapeNet Part dataset to ModelNet40 dataset.}
\vspace{-0.1in}
\begin{center}
\begin{tabular}{lcc}
\hline
\textbf{Method} & \textbf{Year} & \textbf{Accuracy} (\%) \\
\hline
SPH \cite{2003Rotation} & 2003 & 68.2 \\
LFD \cite{0M} & 2003 & 75.5 \\
T-L Network \cite{2016Learning1} & 2016 & 74.4 \\
VConv-DAE \cite{2016VConv} & 2016 & 75.5 \\
3D-GAN \cite{2016Learning} & 2016 & 83.3 \\
Latent GAN \cite{achlioptas2018learning} & 2018 & 85.7 \\
MRTNet-VAE \cite{2018Multiresolution} & 2018 & 86.4 \\
FoldingNet \cite{yang2018foldingnet} & 2018 & 88.4 \\
Contrast-Cluster \cite{zhang2019unsupervised} & 2019 & 86.8 \\
PointCapsNet \cite{zhao20193d} & 2019 & 88.9 \\
Multi-task \cite{hassani2019unsupervised} &  2019 & 89.1 \\
GraphTER \cite{gao2020graphter} & 2020 & 89.1 \\ \hline
Ours &  &\textbf{89.6}  \\
\hline
\end{tabular}
\end{center}
\vspace{-0.2in}
\label{tab:transfer}
\end{table}

% Here $\N_{k}\sim q^{-}(\N|\M)$, which depends on the choosing of two thresholds of similarity $b_l$ and $b_u$. $f$ represents the cosine similarity function. As for the linear annealing policy, we set $b_l$ and $b_u$ to be 0 and 1 early in training. Over many epochs, we linearly decrease $b_l$ and $b_u$ to make them approach each other and finally set them as a fixed sub interval of $\left[0, 1\right]$.

% For the choosing positive and negative samples, we use encoder $E_2$ to project them into a feature space $\mathcal{Z} \in \mathbf{R}^{d_z}$ first, where $d_z \gg d_s$. Higher dimension can lead to encode richer local information. Next, we put the feature vectors into the GCN layer $\mathcal{A}$ to aggregate the local feature and get the global feature. At last, the cosine similarity of feature vectors of positive samples and negative samples are calculated to calculate the loss.
\vspace{-0.1in}
\subsection{Analysis}
We further analyze the connection of our method with the recently developed self-attention \cite{zhao2020point} and point cloud transformer \cite{guo2020pct}.
Self-attention is an attention mechanism relating different positions of spatial and temporal data to model the context dependency within the data, which has shown its effectiveness in machine reading, language understanding, image description generation, and so on.
The well-known transformer models \cite{khan2021transformers} heavily rely on self-attention mechanism to exploit global or long-range dependencies between the input and output.
In essence, the nonlocal self-similarity exploited in our self-contrastive learning model is consistent with the principle of self-attention.
As discussed in \cite{wang2018non}, self-attention is a special case of non-local operations in the embedded Gaussian version.

On the other hand, the nonlocal self-similarity of a point cloud distinguishes from the commonly used self-attention or transformer in the explicit geometric meaning as well as the extension to positive and negative sampling in contrastive learning.
Moreover, the proposed model samples self-similar patches as positives, while patches with less degree of similarity are treated as hard negatives.
By relating and contrasting patches in different locations of a point cloud, our model captures both local geometric details and nonlocal semantic primitives, leading to accurate local and global representations.
This is beneficial to downstream tasks such as point cloud segmentation, which will be evaluated in the experiments.

\begin{table*}[t]
\centering
\small
\caption{Part segmentation results on ShapeNet Part dataset. Metric is mIoU (\%) on points. ``Sup.'' denotes \textit{supervised}.}
\label{tab:segmentation}
\vspace{-0.1in}
\begin{tabular}{r|p{0.65cm}<{\centering}|p{0.65cm}<{\centering}|p{0.45cm}<{\centering}p{0.45cm}<{\centering}p{0.45cm}<{\centering}p{0.45cm}<{\centering}p{0.45cm}<{\centering}p{0.45cm}<{\centering}p{0.45cm}<{\centering}p{0.45cm}<{\centering}p{0.45cm}<{\centering}p{0.45cm}<{\centering}p{0.45cm}<{\centering}p{0.45cm}<{\centering}p{0.45cm}<{\centering}p{0.45cm}<{\centering}p{0.45cm}<{\centering}p{0.45cm}<{\centering}}
% \toprule
\hline
Method & Sup. & Mean & Aero & Bag & Cap & Car & Chair & \begin{tabular}[c]{@{}c@{}}Ear\\ Phone\end{tabular} & Guitar & Knife & Lamp & Laptop & Motor & Mug & Pistol & Rocket & \begin{tabular}[c]{@{}c@{}}Skate\\ Board\end{tabular} & Table \\ \hline
Samples & & & 2690 & 76 & 55 & 898 & 3758 & 69 & 787 & 392 & 1547 & 451 & 202 & 184 & 283 & 66 & 152 & 5271 \\ \hline
PointNet \cite{qi2017pointnet} & $\checkmark$ & 83.7 & 83.4 & 78.7 & 82.5 & 74.9 & 89.6 & 73.0 & 91.5 & 85.9 & 80.8 & 95.3 & 65.2 & 93.0 & 81.2 & 57.9 & 72.8 & 80.6 \\
PointNet++ \cite{qi2017pointnet++} & $\checkmark$ & 85.1 & 82.4 & 79.0 & 87.7 & 77.3 & 90.8 & 71.8 & 91.0 & 85.9 & 83.7 & 95.3 & 71.6 & 94.1 & 81.3 & 58.7 & 76.4 & 82.6 \\
KD-Net \cite{klokov2017escape} & $\checkmark$ & 82.3 & 80.1 & 74.6 & 74.3 & 70.3 & 88.6 & 73.5 & 90.2 & 87.2 & 81.0 & 94.9 & 57.4 & 86.7 & 78.1 & 51.8 & 69.9 & 80.3 \\
SynSpecCNN \cite{yi2017syncspeccnn} & $\checkmark$ & 84.7 & 81.6 & 81.7 & 81.9 & 75.2 & 90.2 & 74.9 & 93.0 & 86.1 & 84.7 & 95.6 & 66.7 & 92.7 & 81.6 & 60.6 & 82.9 & 82.1 \\
PCNN \cite{atzmon2018point} & $\checkmark$ & 85.1 & 82.4 & 80.1 & 85.5 & 79.5 & 90.8 & 73.2 & 91.3 & 86.0 & 85.0 & 95.7 & 73.2 & 94.8 & 83.3 & 51.0 & 75.0 & 81.8 \\
PointCNN \cite{li2018pointcnn} & $\checkmark$ & 86.1 & 84.1 & 86.5 & 86.0 & 80.8 & 90.6 & 79.7 & 92.3 & 88.4 & 85.3 & 96.1 & 77.2 & 95.3 & 84.2 & 64.2 & 80.0 & 83.0 \\
DGCNN \cite{wang2019dynamic} & $\checkmark$ & 85.2 & 84.0 & 83.4 & 86.7 & 77.8 & 90.6 & 74.7 & 91.2 & 87.5 & 82.8 & 95.7 & 66.3 & 94.9 & 81.1 & 63.5 & 74.5 & 82.6 \\
RS-CNN \cite{liu2019relation} & $\checkmark$ & 86.2 & 83.5 & 84.8 & 88.8 & 79.6 & 91.2 & 81.1 & 91.6 & 88.4 & 86.0 & 96.0 & 73.7 & 94.1 & 83.4 & 60.5 & 77.7 & 83.6\\ \hline
%RS \cite{sauder2019self} & 85.3 & 84.1 & 84.0 & 85.8 & 77.0 & 90.9 & 80.0 & 91.5 & 87.0 & 83.2 & 95.8 & 71.6 & 94.0 & 82.6 & 60.0 & 77.9 & 81.8 \\
LGAN \cite{achlioptas2018learning} & $\times$ & 57.0 & 54.1 & 48.7 & 62.6 & 43.2 & 68.4 & 58.3 & 74.3 & 68.4 & 53.4 & 82.6 & 18.6 & 75.1 & 54.7 & 37.2 & 46.7 & 66.4 \\
MAP-VAE \cite{han2019multi} & $\times$ & 68.0 & 62.7 & 67.1 & \textbf{73.0} & 58.5 & 77.1 & \textbf{67.3} & 84.8 & 77.1 & 60.9 & 90.8 & 35.8 & 87.7 & 64.2 & 45.0 & 60.4 & \textbf{74.8} \\
GraphTER (1 FC) \cite{gao2020graphter} & $\times$ & 62.5 & 55.7 & 44.8 & 36.5 & 18.9 & 78.4 & 26.3 & 76.2 & 76.8 & 51.9 & 89.0 & 12.0 & 47.9 & 55.9 & 22.7 & 32.9 & 70.8 \\
\textbf{Ours (1 FC)} & $\times$ & \textbf{76.0} & \textbf{71.8} & \textbf{71.2} & 70.1 & \textbf{59.4} & \textbf{83.1} & 56.1 & \textbf{86.6} & \textbf{77.8} & \textbf{73.9} & \textbf{93.4} & \textbf{52.5} & \textbf{89.8} & \textbf{73.7} & \textbf{47.3} & \textbf{69.9} & 73.4 \\
\hline
GraphTER (5 FCs) \cite{gao2020graphter} & $\times$ & 81.9 & 81.7 & 68.1 & \textbf{83.7} & 74.6 & \textbf{88.1} & 68.9 & \textbf{90.6} & \textbf{86.6} & \textbf{80.0} & \textbf{95.6} & 56.3 & 90.0 & 80.8 & 55.2 & 70.7 & 79.1 \\
\textbf{Ours (5 FCs)} & $\times$ & \textbf{82.3} & \textbf{82.1} & \textbf{74.5} & 83.6 & \textbf{74.9} & 87.9 & \textbf{72.4} & 89.9 & 85.4 & 79.1 & 95.2 & \textbf{67.3} & \textbf{93.3} & \textbf{81.0} & \textbf{58.2} & \textbf{74.0} & \textbf{79.2} \\ \hline
% \bottomrule
\end{tabular}
\vspace{-0.1in}
\label{tab:unsup_seg}
\end{table*}

\section{Experimental Results}
In this section, we first validate the feature representations of point clouds learned with our model in a transfer learning setting \cite{sauder2019self} in Sec.~\ref{subsec:transfer}.
Then in Sec.~\ref{subsec:seg},
We compare the proposed part segmentation method with state-of-the-art supervised and unsupervised approaches. In our experiments, we employ ModelNet40 \cite{wu20153d} and ShapeNet Part \cite{yi2016scalable} datasets to evaluate the generalizability of our model in a transfer learning strategy.

\textbf{ModelNet40.}
This dataset contains $12,311$ models from $40$ categories, where $9,843$ models are used for training and $2,468$ models are for testing.
We sample $1,024$ points from the original model.

\textbf{ShapeNet Part.}
This dataset contains $16,881$ point clouds from $16$ object categories, annotated with $50$ different parts.
We sample $2,048$ points from each 3D point cloud.
We employ $12,137$ models for training, and $2,874$ models for testing.

\subsection{Transfer learning}
\label{subsec:transfer}

\subsubsection{Implementation Details}
In this task, the similarity-learning network and the contrastive network are both trained via the Adam optimizer \cite{kingma2014adam} with a batch size of $32$ and an initial learning rate of $0.002$. The similarity-learning network is trained for 128 epochs and the contrastive learning network is trained for 512 epochs.
The learning rates of the two networks are scheduled to decrease by 0.8 every $20$ epochs and $0.5$ every $50$ epochs, respectively.

For the similarity-learning network, we deploy four EdgeConv \cite{wang2019dynamic} layers and one fully-connected layer as the encoder $E_1(\cdot)$.
The number of nearest neighbors $k$ is set to 20 for all EdgeConv layers.
After the four EdgeConv layers, the 128-dimensional point-wise feature matrix goes through a fully-connected layer with input channels of $128$ and output channels of $32$.
We then sample similar and dissimilar patch pairs as described in Sec.~\ref{subsec:self_similar_train}, and the point-wise features of these patches are fed into an aggregation operator $G_1(\cdot)$, which consists of a GCN \cite{kipf2017semi} and average pooling layer to acquire the representations of patches.
The representations of similar and dissimilar patch pairs are first concatenated to form a $64$-dimensional feature vector, and then fed into a discriminator $D(\cdot)$ containing one fully-connected layer to predict the similarity of each pair.

In the linear annealing strategy for hard negative sampling, given that most similarity values between different patch pairs range from 0.8 to 1.0, we opt to reduce the value of $b_u$ by 0.025 and increase the value of $b_l$ by 0.05 for every 20 epochs after 300 training epochs of using the whole negative samples.
In the encoder $E_2(\cdot)$ of the contrastive network, we adopt eight EdgeConv layers with the number of nearest neighbors $k=20$ to acquire the point-wise feature representations.
We concatenate the point-wise features from the eight EdgeConv layers, which will be fed into a fully-connected layer to output a 256-dimensional point-wise feature matrix.
Similar to the similarity-learning network, we also use a GCN layer and average pooling layer as our aggregation operator $G_2(\cdot)$ to aggregate the feature representations of each patch.
Besides, the batch normalization layer and LeakyReLU activation function with a negative slope of 0.2 are employed after each convolutional layer.

During the evaluation stage, the weights of the encoder $E_2(\cdot)$ are fixed to extract the point-wise feature representations of point clouds.
Then an average pooling layer is deployed to acquire the global features, after which a linear SVM classifier is trained to map the global features to classification scores.

\begin{figure}[t]
  \centering
  \includegraphics[width=0.9\columnwidth]{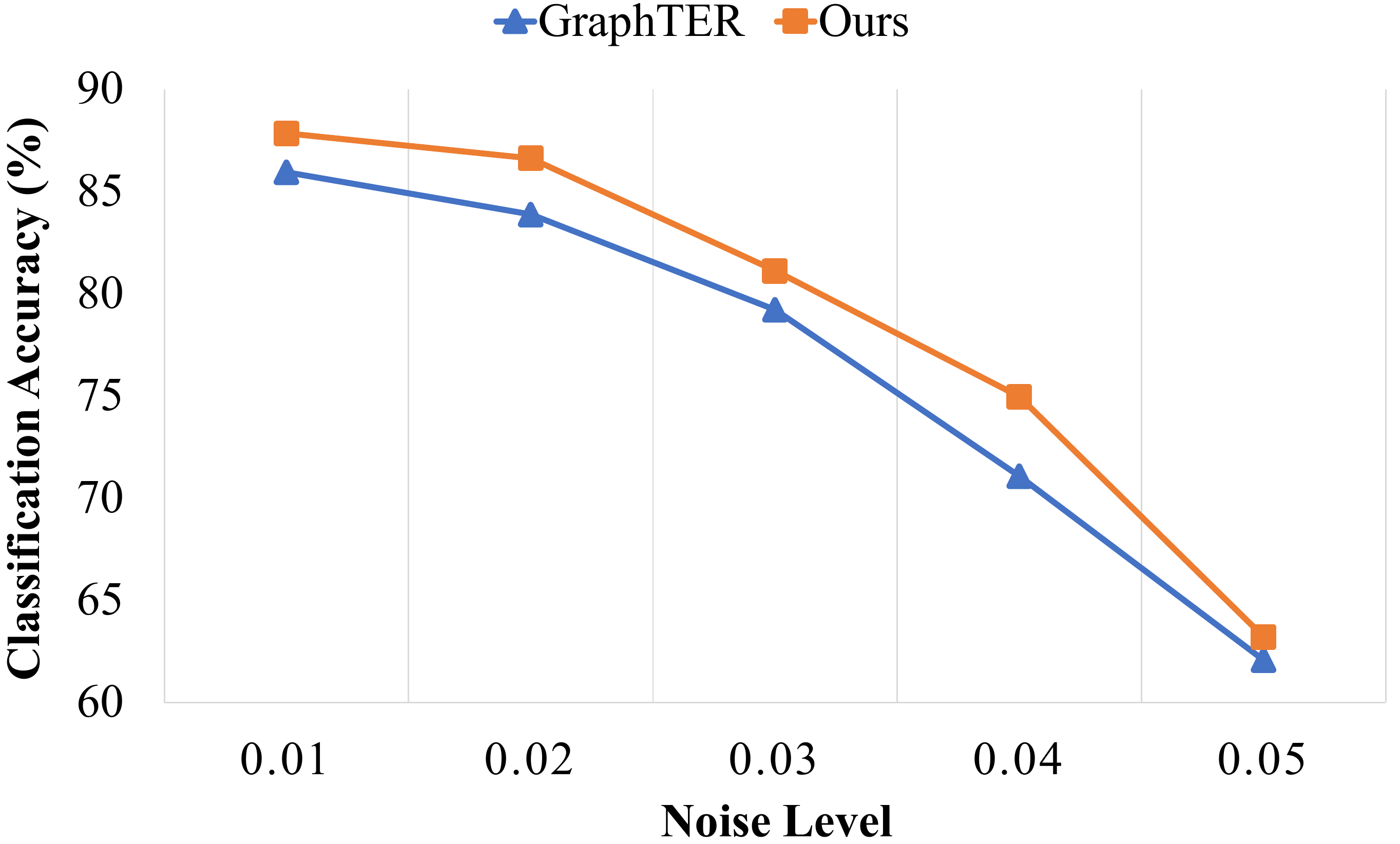}
  \vspace{-0.1in}
  \caption{Comparison in classification accuracy with the most competitive method GraphTER on ModelNet40 dataset at different Gaussian noise levels.}
  \vspace{-0.2in}
  \label{fig:noise_level_cls}
\end{figure}
\subsubsection{Classification Results}
%In this section, we show the efficiency of the proposed method in 3D point cloud representation learning.
We follow the same setting in \cite{zhang2019unsupervised} to train a linear SVM classifier on the feature representations of ModelNet40 obtained from the contrastive network. Both similarity-learning and contrastive networks are trained on the ShapeNet part dataset.
As shown in Tab.~\ref{tab:transfer}, the proposed method outperforms all other unsupervised competitive methods on the ModelNet40 dataset, which justifies the effectiveness of our method.

\begin{figure}[t]
  \centering
  \includegraphics[width=0.9\columnwidth]{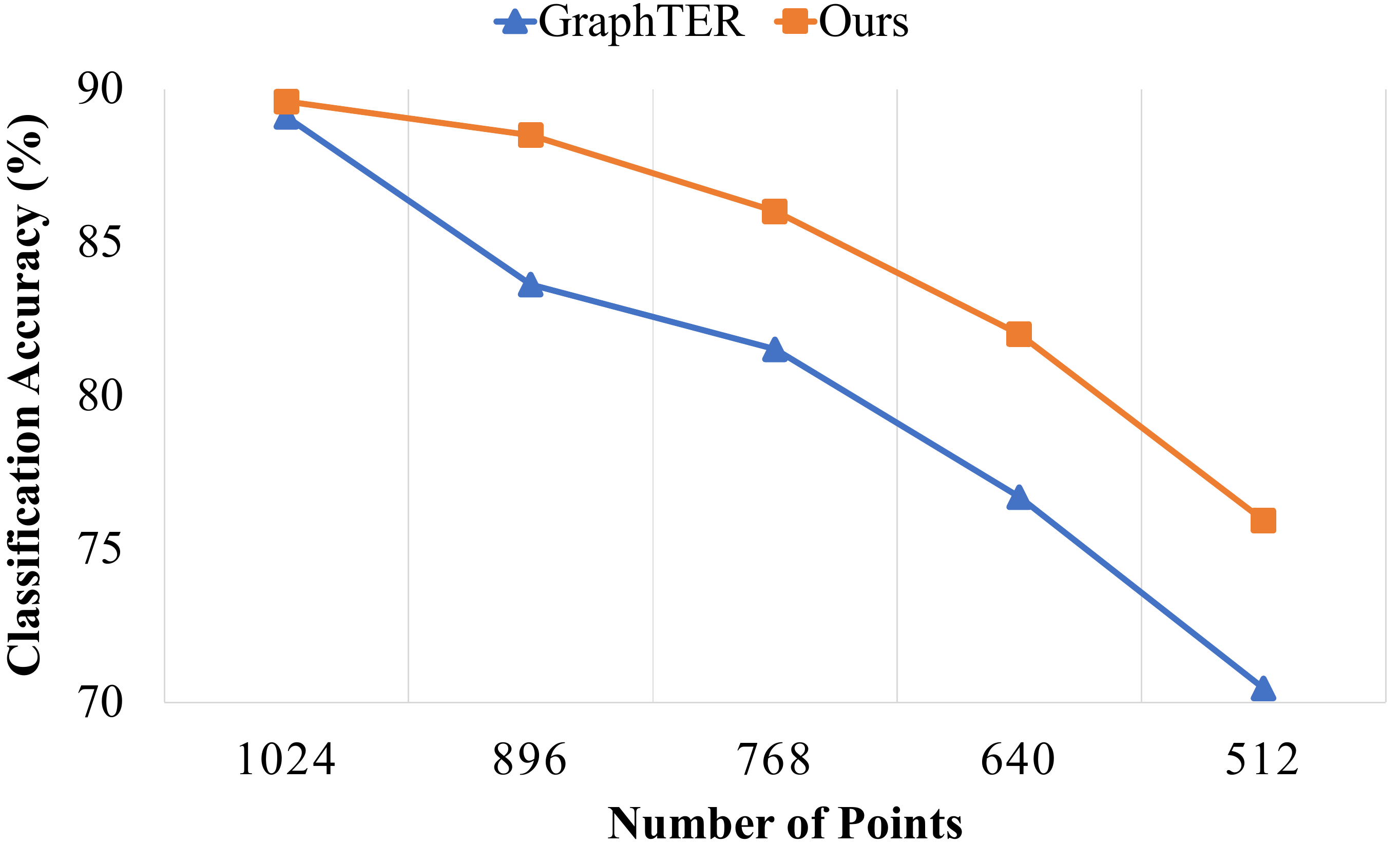}
  \vspace{-0.1in}
  \caption{Comparison in classification accuracy with the most competitive method GraphTER on ModelNet40 dataset at different point cloud densities.}
  \vspace{-0.2in}
  \label{fig:density}
\end{figure}

\subsubsection{Ablation Studies}
We test the robustness of our model at different noise levels and densities, which is important to real-world applications where point cloud data often suffer from noise or low density.

\textbf{Robustness to Noise.} In order to test the robustness of our model to random noise, we randomly jitter the original coordinates of 3D point clouds with Gaussian noise, with zero mean and standard deviation $\sigma \in \{0.01,0.02,0.03,0.04,0.05\}$.
As presented in Fig.~\ref{fig:noise_level_cls}, we compare our method with the most competitive method GraphTER \cite{gao2020graphter} on ModelNet40 dataset, where the horizontal axis denotes the noise level (standard deviation $\sigma$), and the vertical axis is the classification accuracy evaluated from a linear SVM classifier.
We observe that the proposed model is more robust, with the classification accuracy of $63.2\%$ even when the noise level $\sigma=0.05$.
We also see that the performance trend of GraphTER is similar and comparable to ours. This is because GraphTER employs point-wise transformation (translation, rotation or shearing) for data augmentation, which is beneficial to the robustness of the model to slight noise. By contrast, we do not need this strategy but still achieve satisfactory classification results and good robustness.

% \begin{table}[]
% \centering
% \caption{accuracies on different noise levels.}
% \begin{tabular}{|l|c|c|c|c|c|c|c|}
% \hline
% \multicolumn{1}{|c|}{Noise Level} & 0.01 & 0.02 & 0.03& 0.04 &0.05 \\ \hline
% GraphTER & 85.94 & 83.87 & 79.21 & 71.07 & 62.07  \\ \hline
% ours & 87.84 & 86.62 & 81.12 & 74.96 & 63.21   \\ \hline
% % {Noise Level} & 0.002 & 0.004 & 0.006 & 0.008 & 0.010 \\ \hline
% % ours & 88.70&  88.41& 88.05 &87.93  & 87.84 \\ \hline
% \end{tabular}
% \label{tab:noise_level}
% \end{table}

\textbf{Robustness to Density.} We further test the robustness of our model to 3D point clouds with low density.
We randomly sample $\{1024,896,768,640,512\}$ points from the original point cloud on ModelNet40 dataset, and compare our method with GraphTER \cite{gao2020graphter}.
As shown in Fig.~\ref{fig:density}, the classification accuracy of our model keeps $75.9\%$ when the number of points is $512$, which outperforms GraphTER ($70.5\%$) significantly.
Hence, the proposed method is robust to noisy and sparse point clouds.
This gives credit to the proposed contrastive learning based on nonlocal self-similarity, which captures the relationship between parts of the point cloud.

% \begin{table}[]
% \centering
% \caption{accuracies on different densities.}
% \label{tab:num_points}
% \begin{tabular}{|l|c|c|c|c|}
% \hline
% \multicolumn{1}{|c|}{\#Points} & 896 & 768 & 640 & 512 \\ \hline
% GraphTER & 83.63 & 81.52 & 76.70 & 70.46  \\ \hline
% Ours & 88.50 & 86.02 & 82.01 & 75.93  \\ \hline
% \end{tabular}
% \end{table}

\subsection{Part Segmentation}
\label{subsec:seg}

\subsubsection{Implementation Details}

We also use Adam optimizer to train the similarity-learning network and the contrastive network for 128 epochs and 512 epochs, respectively.
The hyper-parameters are the same as in Sec.~\ref{subsec:transfer}.

We adopt a linear classifier with several fully-connected layers to map the point-wise feature representations to segmentation scores, with the input of a 256-dimensional feature vector from the EdgeConv layer of the contrastive network.
We use the negative log-likelihood loss to train the classifier.

% The contrastive model is implemented with Pytorch on ... with ... GPUs. ... We distribute the training mini-batches over all GPUs and average the gradients to update the model parameters. With this setting, our contrastive model takes 1240s on average to train one epoch on the ShapeNet.

% In particular, we sample one point cloud for training if the number of point clouds with some part multiplied by the sampling rate is less than 1. The results show that we can still achieve the segmentation effect of reasonable even under the extreme condition of minimal training data.

% For the unsupervised setting, we adopt a 1-layer MLP to serve as our classifier for part segmentation to verify that the segmentation effect of our model does not depend on the downstream classifier. In addition, we also apply a 5-layer MLP as our classifier like \cite{gao2020graphter}. During the whole training procedule, the weights of the encoder of the contrastive model are kept frozen.

\subsubsection{Unsupervised Results}

We adopt the Intersection-over-Union (IoU) metric and follow the same evaluation protocol as in the PointNet \cite{qi2017pointnet}: the IoU of a shape is computed by averaging the IoUs of all shapes belonging to that category. The mean IoU (mIoU) is finally calculated by averaging the IoUs of all test shapes.

We compare our model with both unsupervised approaches and supervised approaches, as listed in Tab.~\ref{tab:unsup_seg}.
For fair comparison, we use a different number of fully-connected layers as our classifier to compare with other methods: 1) one fully-connected layer (denoted as ``\textbf{1 FC}'') as in \cite{han2019multi}, and 2) five fully-connected layers (denoted as ``\textbf{5 FCs}'') as in \cite{gao2020graphter}.
Note that we reproduce the segmentation results of GraphTER under the ``1 FC'' setting.
Under the challenging ``1 FC'' setting, we achieve an mIoU of $76.0\%$, which significantly outperforms the state-of-the-art unsupervised method MAP-VAE \cite{han2019multi} by $8.0\%$, and GraphTER \cite{gao2020graphter} by $13.5\%$.
Under the ``5 FCs'' setting, we achieve an mIoU of $82.3\%$, which also outperforms the state-of-the-art unsupervised method GraphTER \cite{gao2020graphter} by $0.4\%$.
Moreover, the proposed model achieves comparable performance to the state-of-the-art fully supervised approaches, which pushes greatly closer towards the upper bound set by the fully supervised counterparts.

We also observe that our model achieves the top performance in most categories such as \texttt{aeroplane}, \texttt{motorbike} and \texttt{skateboard}, which exhibit strong self-similarity and can be well captured by our self-contrastive learning.
% Other models such as \texttt{Cap} and \texttt{Table} have simple structures and clear nonlocal self-similarity.
% We believe that our model collapses on these categories when there has clear nonlocal self-similarity between patches, and then lead to bad performance.

% In 16 categories and 50 parts of the ShapeNet dataset, our performance have improved a lot in airplanes, mugs and motors.
% Actually, we think it's because of their significant nonlocal self-similarity property. Take planes for example, the wings distributed on the left and right sides are mirrored in space, with their semantic and structural information remaining high self-similarity. The same is true for the tail and engine. However, Our performance on several other categories is not satisfactory, even have negative growth such as on Knifes and lamps. Actually, The self-similarity of these objects like knifes is not obvious, which we think is the main reason for the effect.

\subsubsection{Semi-supervised Results}

\begin{table}[t]
\small
\caption{Segmentation results using different percentages of training data.}
\vspace{-0.1in}
\begin{center}
\begin{tabular}{r|c|c}
\hline
\textbf{Method} & \textbf{5\% of training data}  &  \textbf{1\% of training data} \\
\hline
SO-Net \cite{2018SO} & 69.0 & 64.0\\
PointCapsNet \cite{zhao20193d} & 70.0 & 67.0\\
MortonNet \cite{DBLP:journals/corr/abs-1904-00230} & 77.1 & - \\
Multi-task \cite{hassani2019unsupervised} & 77.7 & 68.2\\
ACD (w/ fine-tune) \cite{gadelha2020labelefficient} & 79.7 & 75.7\\ \hline
Ours (w/o fine-tune) & \textbf{78.2} & \textbf{74.8}\\
Ours (w/ fine-tune) & \textbf{79.2} & \textbf{76.2}\\
\hline
\end{tabular}
\end{center}
\vspace{-0.1in}
\label{tab:semi_seg}
\end{table}

We follow the same setting in \cite{zhao20193d} to test our model with limited labeled training data.
Specifically, we train the linear classifier with only $1\%$ and $5\%$ labeled training data, and test the model on all available test data.
For fair comparison with other methods as presented in Tab.~\ref{tab:semi_seg}, we adopt two settings: 1) Train the linear classifier with weights of the feature extractor \textbf{not frozen}, as in ACD \cite{gadelha2020labelefficient}, denoted as ``\textbf{Ours (w/ fine-tune)}''; 2) Train the linear classifier with weights of the feature extractor \textbf{frozen}, as in other methods, denoted as ``\textbf{Ours (w/o fine-tune)}''.

As listed in Tab.~\ref{tab:semi_seg}, under the ``w/o fine-tune'' setting, our model achieves the state of the art performance, leading to an mIoU of $78.2\%$ with $5\%$ of training data; with $1\%$ of training data, our model still achieves the best performance, with an mIoU of $74.8\%$ that outperforms Multi-task \cite{hassani2019unsupervised} by $8.0\%$.
Under the ``w/ fine-tune'' setting, our result is comparable to ACD \cite{gadelha2020labelefficient} with $5\%$ of training data, and outperforms ACD with $1\%$ of training data.

\begin{figure}[t]
  \centering
  \includegraphics[width=\columnwidth]{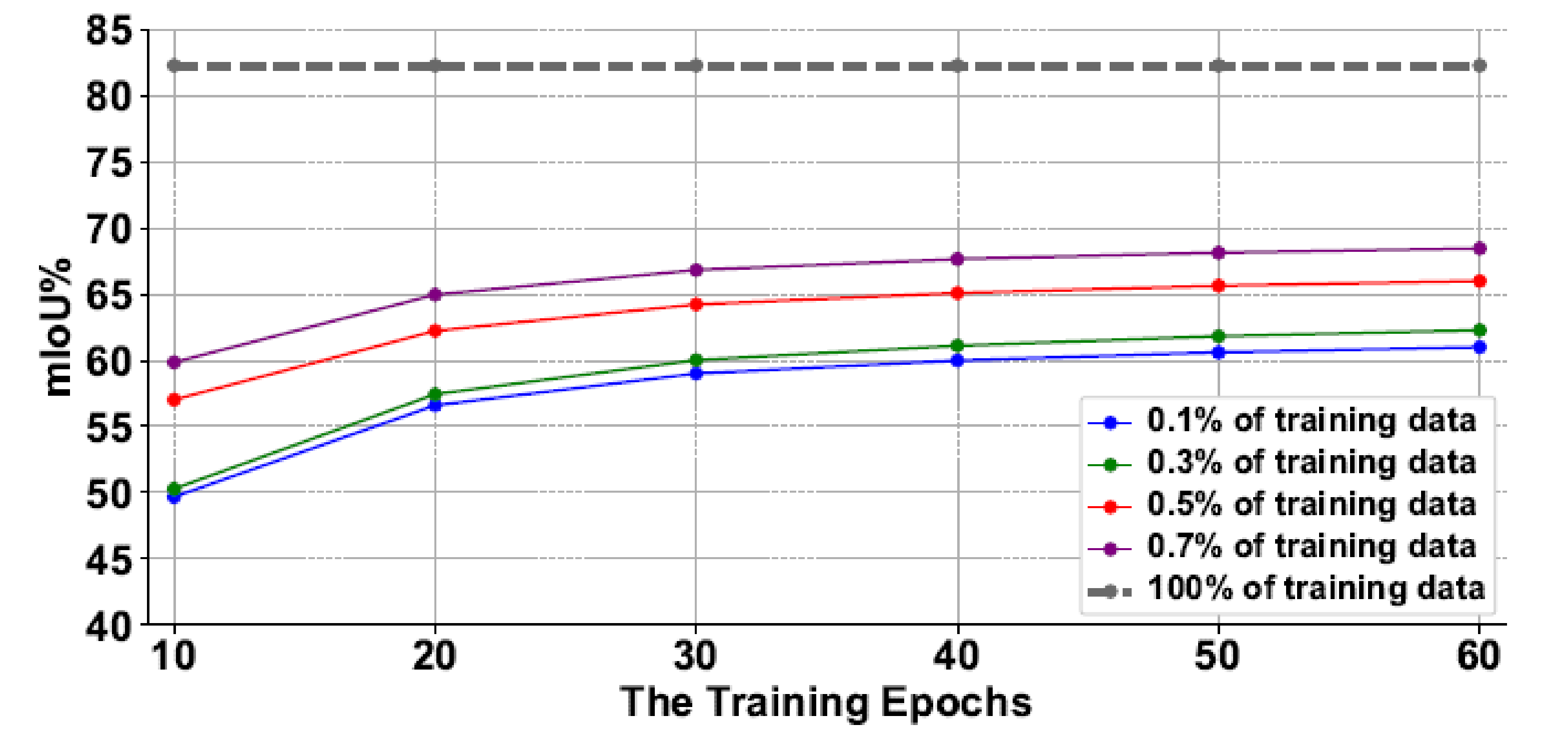}
  \caption{Segmentation results on ShapeNet Part dataset under different percentages of training data.}
  \vspace{-0.1in}
  \label{fig:label_rates}
\end{figure}

We further reduce the amount of training data to evaluate our model.
We adopt four percentages of available training data in $\{0.1\%,0.3\%,0.5\%,0.7\%\}$ to train the linear classifier, with the weights of the feature extractor frozen.
As presented in Fig.~\ref{fig:label_rates}, the horizontal axis denotes the training epoch of the classifier and the vertical axis is the mIoU.
We observe that the mIoU increases with the training epoch and basically converges at the $60$th epoch, which is stable.
The mIoU also increases with the percentage of training data.
At the same time, our model still achieves a reasonable mIoU of $61.11\%$ with $0.1\%$ of training data.
This validates the effectiveness of our model with extremely few labeled training samples.
% , and greatly closes the gap of the upper bounded performance by the $100\%$ of training samples.

\subsubsection{Visualization Results}

\begin{figure}[t]
  \centering
  \subfigure[MAP-VAE]{
  \includegraphics[width=0.4\columnwidth]{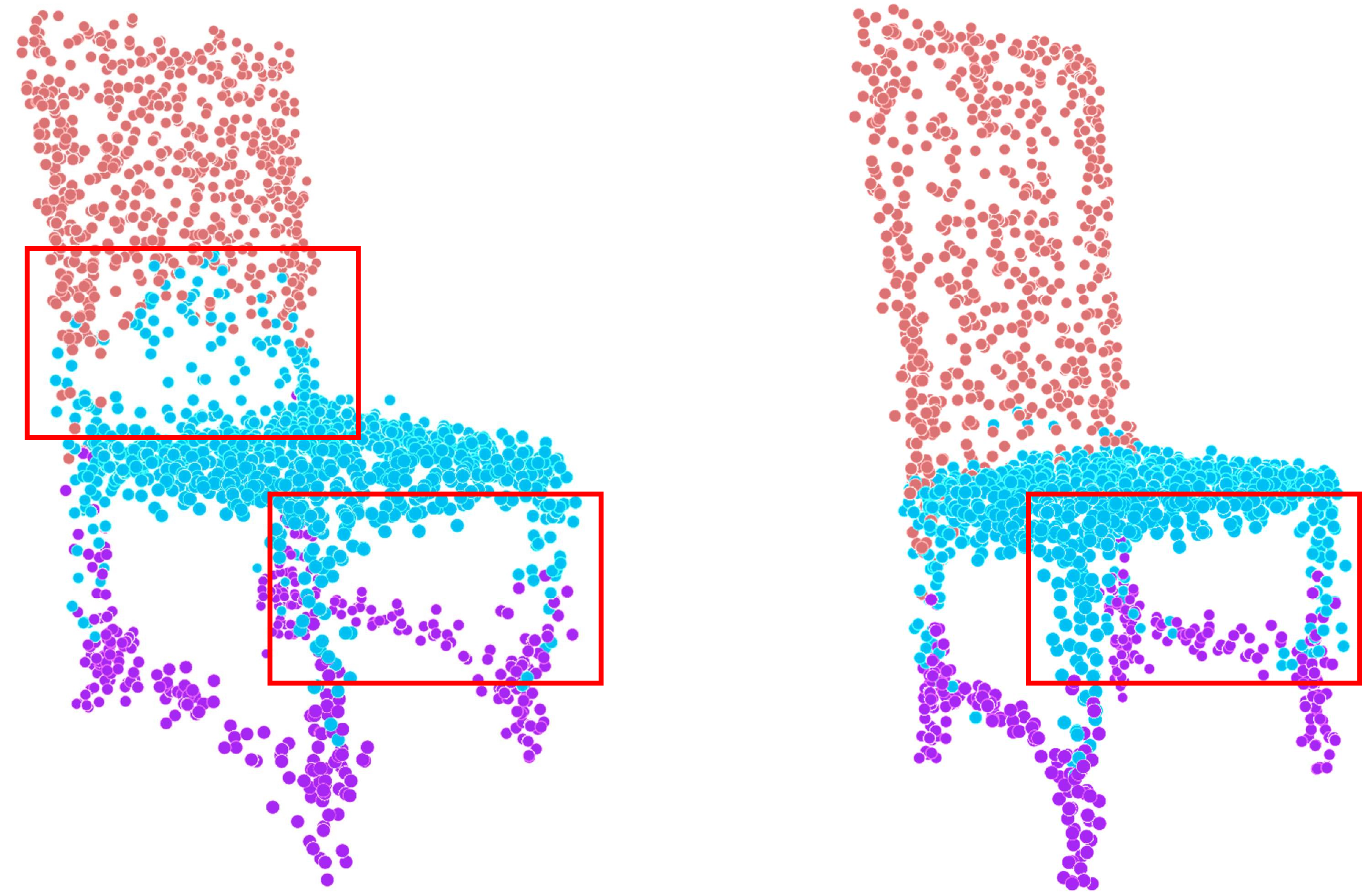}
  \label{subfig:mapvae_chair}
  }
  \hspace{0.15in}
  \subfigure[Ours]{
  \includegraphics[width=0.4\columnwidth]{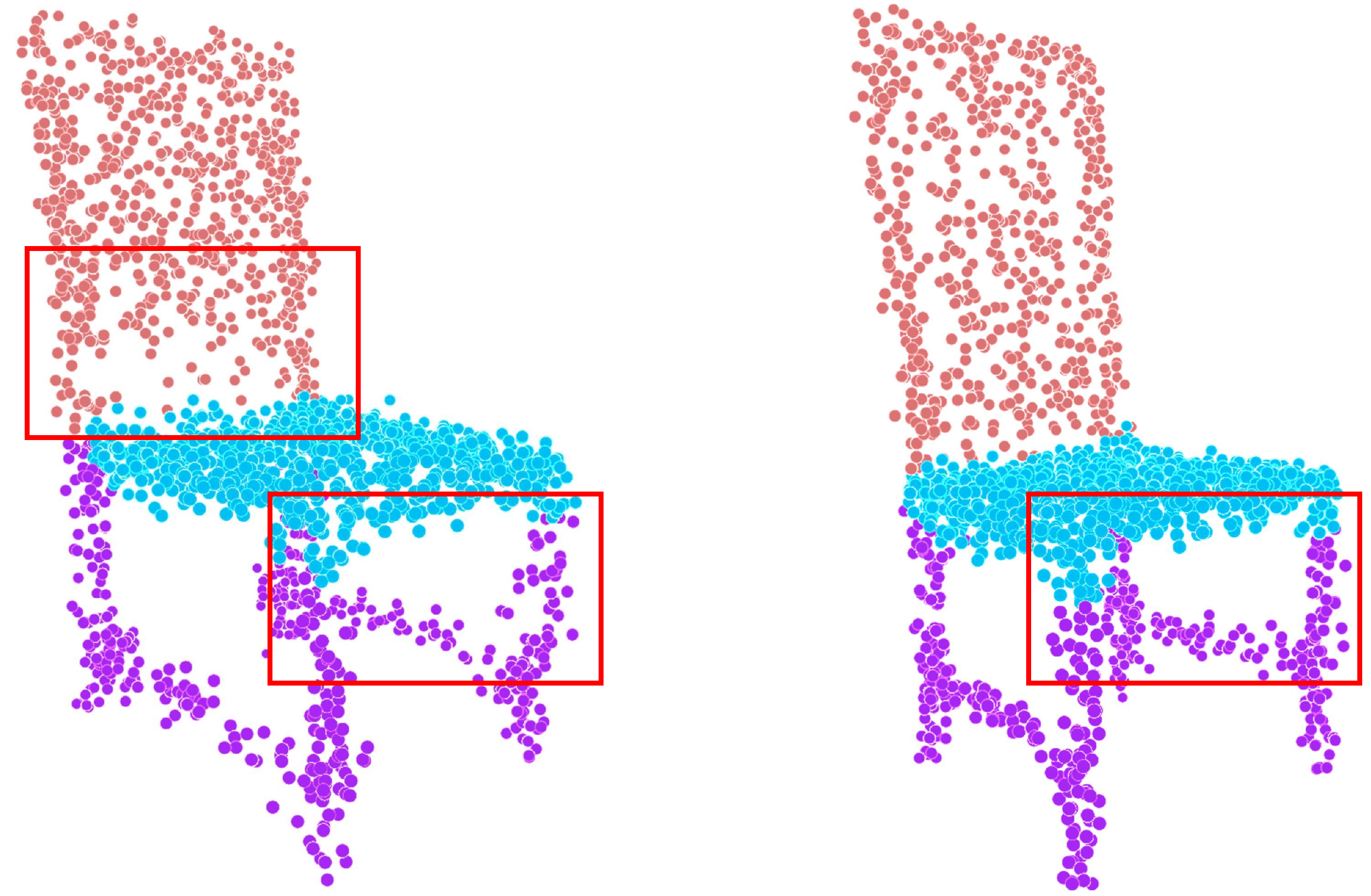}
  \label{subfig:ours_chair}
  }
  \vspace{-0.1in}
  \caption{Visual comparison of part segmentation under the ``1 FC'' setting.}
  \vspace{-0.2in}
  \label{fig:vis_1fc}
\end{figure}

\begin{figure}[t]
  \centering
  \subfigure[Ground-truth]{
  \includegraphics[width=0.3\columnwidth]{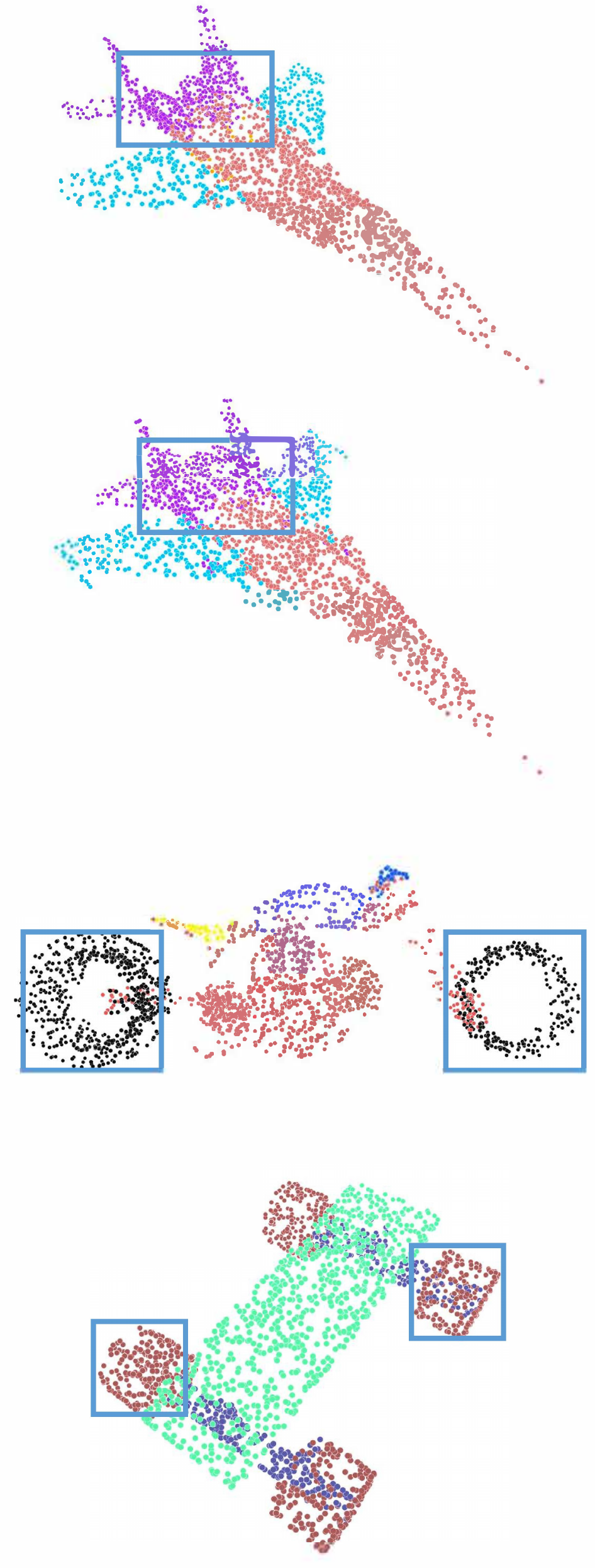}
  \label{subfig:gt_5fc}
  }
  \subfigure[GraphTER]{
  \includegraphics[width=0.3\columnwidth]{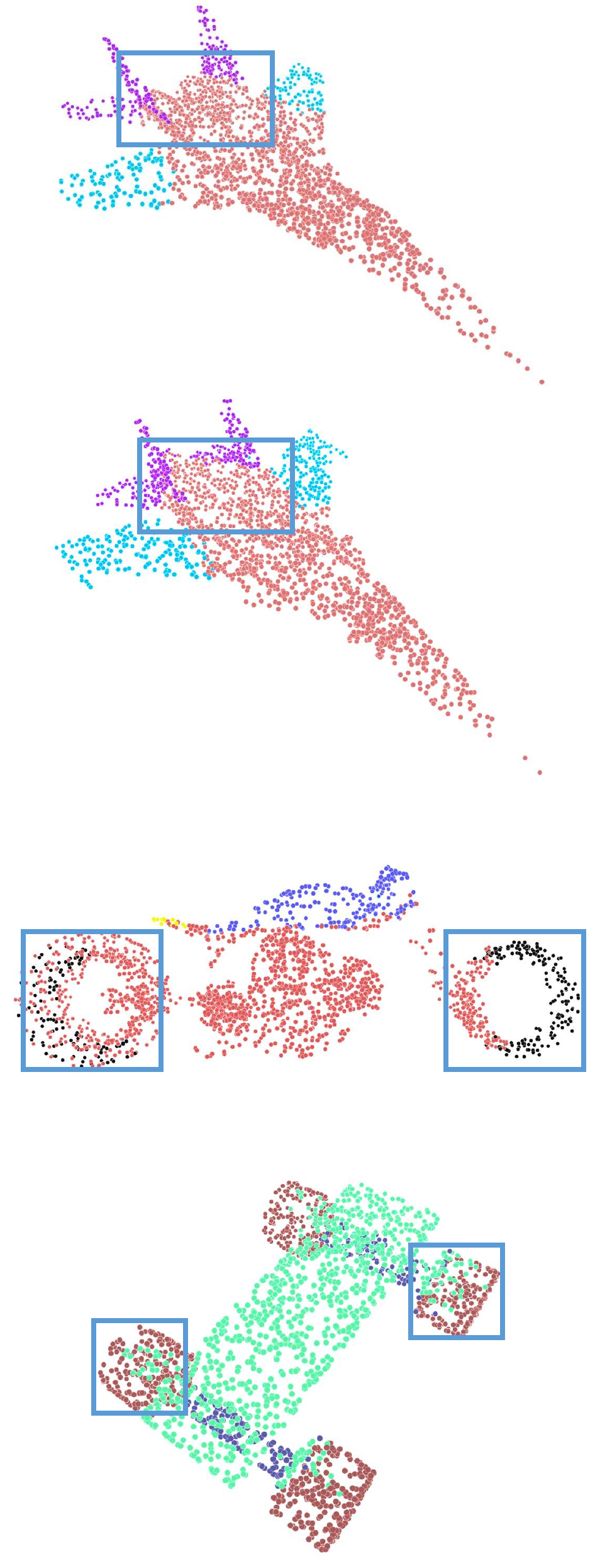}
  \label{subfig:graph_ter_5fc}
  }
  \subfigure[Ours]{
  \includegraphics[width=0.3\columnwidth]{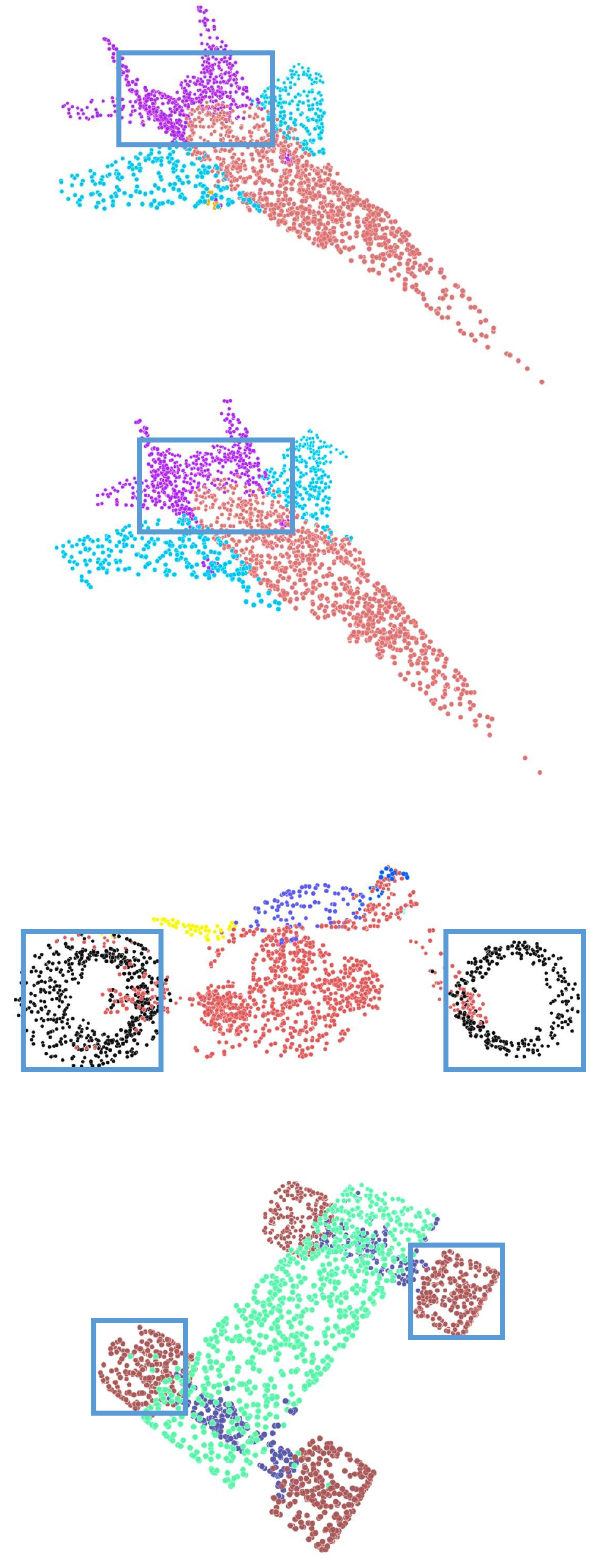}
  \label{subfig:ours_5fc}
  }
  \vspace{-0.1in}
  \caption{Visual comparison of part segmentation under the ``5 FCs'' setting.}
  \vspace{-0.2in}
  \label{fig:vis_5fc}
\end{figure}

Further, we qualitatively compare the proposed method with MAP-VAE \cite{han2019multi} on the \texttt{Chair} model under the ``1 FC'' setting, as illustrated in Fig.~\ref{fig:vis_1fc}.
The proposed model leads to much more accurate segmentation results than MAP-VAE especially in transition areas, \eg, around the legs and the back of the chair.
We also compare our method with the state-of-the-art unsupervised method GraphTER \cite{gao2020graphter} on various models under the ``5 FCs'' setting, as shown in Fig.~\ref{fig:vis_5fc}.
We see that our model produces accurate results in detailed regions, \eg, the tail of the \texttt{aeroplane} model, and the wheels of the \texttt{motorbike} and \texttt{skateboard} models, which benefits from the exploitation of nonlocal similarity.

\subsubsection{Ablation Studies} We evaluate the advantages of hard negative sampling, and test the robustness of our model.

\textbf{Advantages of Hard Negative Sampling.} To validate the advantages of the proposed hard negative sampling, we simply take all other patches as negative samples to the anchor patch.
As presented in Tab.~\ref{tab:without_hard_neg}, we see that with hard negative sampling, the performance of our model outperforms that of our model without hard negative sampling by a large margin in all experimental settings.
Specifically, our model improves the mIoU by an average of $6.35\%$ in unsupervised setting and $5.5\%$ in semi-supervised setting over the model without hard negative sampling, proving that the proposed hard negative sampling makes significant contributions to our method for point cloud representation learning.

\begin{table}[t]
\centering
\small
\caption{Ablation studies on the proposed Hard Negative Sampling in terms of segmentation mIoU (\%) on ShapeNet Part.}
\vspace{-0.1in}
\begin{tabular}{r|r|p{1.3cm}<{\centering}|c}
\hline
\multicolumn{2}{c|}{\multirow{2}{*}{\textbf{Setting}}} & \multicolumn{2}{c}{\textbf{Hard Negative Sampling}} \\ \cline{3-4}
\multicolumn{2}{c|}{} & \textbf{w/} & \textbf{w/o} \\ \hline
\multirow{2}{*}{Unsupervised} & 1 FC & \textbf{76.0} & 68.8 \\ %\cline{2-4}
 & 5 FCs & \textbf{82.3} & 76.8 \\ \hline
\multirow{2}{*}{5\% of training data} & w/ fine-tune & \textbf{79.2} & 75.2 \\ %\cline{2-4}
 & w/o fine-tune & \textbf{78.2} & 72.0 \\ \hline
\multirow{2}{*}{1\% of training data} & w/ fine-tune & \textbf{76.2} & 72.8 \\ %\cline{2-4}
 & w/o fine-tune & \textbf{74.8} & 66.4 \\ \hline
\end{tabular}
\label{tab:without_hard_neg}
\end{table}

\begin{figure}[t]
  \centering
  \includegraphics[width=0.9\columnwidth]{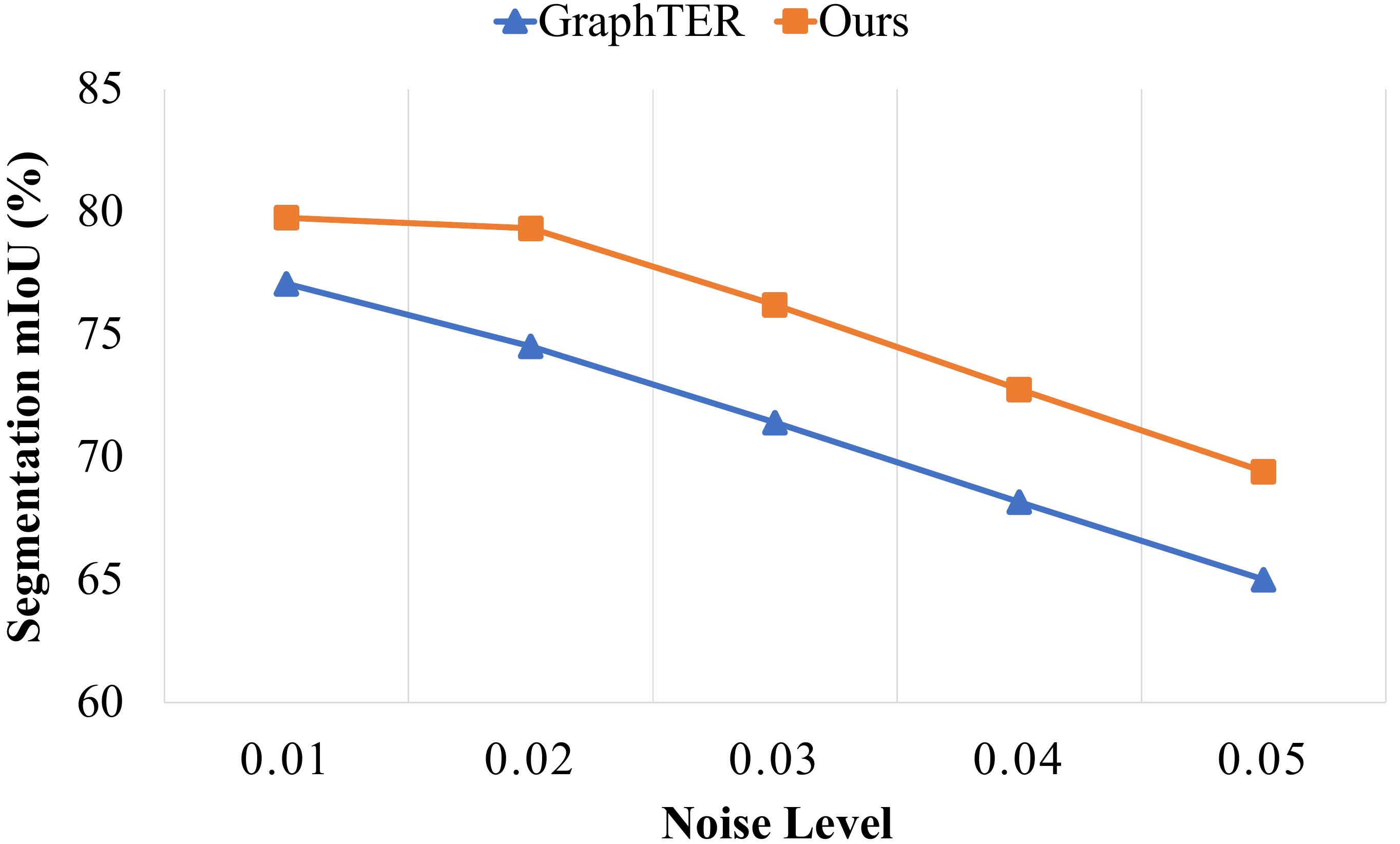}
  \vspace{-0.1in}
  \caption{Comparison in segmentation mIoU with the most competitive method GraphTER on ModelNet40 dataset at different Gaussian noise levels.}
  \vspace{-0.1in}
  \label{fig:noise_level_seg}
\end{figure}

\textbf{Robustness to Noise.} We also test the robustness of our model to noise. We randomly jitter the original coordinates of 3D point clouds with Gaussian noise, with zero mean and standard deviation $\sigma \in \{0.01,0.02,0.03,0.04,0.05\}$.
We compare our model with GraphTER on ShapeNet Part dataset in the ``5 FCs'' setting.
As shown in Fig.~\ref{fig:noise_level_seg}, our model outperforms GraphTER at all the noise levels and achieves reasonable segmentation results even at high noise levels.

% \begin{table}[htbp]
% \small
% \caption{robustness to noise on segmentation.}
% \begin{center}
% \begin{tabular}{|r|c|c|c|c|c|}
% \hline
% Noise level & 0.01 & 0.02 & 0.03 & 0.04 & 0.05 \\
% GraphTER & 77.07 & 74.52 & 71.41 & 68.17 & 65.01 \\ % TO modify
% Ours &  79.76 & 79.33 & 76.21 & 72.76 & 69.41 \\
% \hline
% \end{tabular}
% \end{center}
% \label{tab:semi_seg}
% \end{table}

% \begin{table}[htbp]
% \small
% \caption{Ablation study for segmentation without hard negative sampling.}
% \label{tab:segmentation}
% \begin{center}
% %\begin{tabular}{r|p{0.65cm}<{\centering}|p{0.65cm}<{\centering}|p{0.%65cm}<{\centering}}
% \begin{tabular}{l|c|c}
% Setting & With *** & Without *** \\ \hline
% unsupervised(1fc) & 76.0 & 68.8 \\
% unsupervised(5fc) & 82.3 & 76.8 \\
% semi-supervised(5\% with fine-tune) & 79.2 & 75.2 \\
% semi-supervised(5\% without fine-tune) & 78.2 & 72.0 \\
% semi-supervised(1\% with fine-tune) & 76.2 & 72.8 \\
% semi-supervised(1\% without fine-tune) & 74.8 & 66.4 \\
% \end{tabular}
% \end{center}
% \label{tab:unsup_seg}
% \end{table}

\section{Conclusion}
We propose a self-contrastive learning framework with hard negative sampling based on nonlocal self-similarity, aiming at accurate point cloud representation learning in a self-supervised fashion.
It consists of three building blocks, namely self-similarity training, positive/negative sampling (especially hard negative sampling) from nonlocal self-similarity, and self-contrastive learning with a linear annealing schedule.
By exploiting self-similarity in point cloud, our method has achieved state-of-the-art results in segmentation and transfer learning for classification on popular benchmark datasets. We believe the exploitation of nonlocal self-similarity will also benefit other point cloud tasks such as denoising, reconstruction, and generation.

\clearpage

\balance
\bibliographystyle{ACM-Reference-Format}
%\bibliography{main}
%%% -*-BibTeX-*-
%%% Do NOT edit. File created by BibTeX with style
%%% ACM-Reference-Format-Journals [18-Jan-2012].

\end{document}